\documentclass[twoside,leqno,twocolumn]{article}

\usepackage[letterpaper]{geometry}

\usepackage{ltexpprt}
\usepackage{hyperref}

\usepackage{graphicx}
\usepackage{booktabs}
\usepackage{etoolbox,xspace}
\usepackage{xcolor}
\usepackage{multirow}
\usepackage{paralist}
\usepackage{url}
\usepackage{amsmath}
\usepackage{algorithm}
\usepackage{algpseudocode}
\usepackage{amssymb}
\begin{document}

\newsavebox\CBox
\def\textBF#1{\sbox\CBox{#1}\resizebox{\wd\CBox}{\ht\CBox}{\textbf{#1}}}

\newcommand{\cdash}{\multicolumn{1}{c}{--} }

\newcommand{\super}{$^\blacktriangle$}
\newcommand{\sub}{$^\triangledown$}

\newcommand{\qap}{$\bm{q}_{\textbf{AP}}$}
\newcommand{\qroc}{$\bm{q}_{\textbf{ROC}}$}
\newcommand{\qprn}{$\bm{q}_{\textbf{Prec}}$}

\newcommand{\methodtitle}{{\Large \sf TSA}\xspace}

\newcommand{\method}{{\sf TSAP}\xspace}

\newcommand*{\bord}{\multicolumn{1}{c|}{}}

\newcommand{\cbit}{\begin{compactitem}}
	\newcommand{\ceit}{\end{compactitem}}
\newcommand{\cben}{\begin{compactenum}}
	\newcommand{\ceen}{\end{compactenum}}


\newcommand{\bit}{\begin{itemize}}
	\newcommand{\eit}{\end{itemize}}
\newcommand{\ben}{\begin{enumerate}}
	\newcommand{\een}{\end{enumerate}}
\newcommand{\beq}{\begin{equation}}
\newcommand{\eeq}{\end{equation}}

\newcommand{\ap}{AP\xspace}
\newcommand{\roc}{ROC\xspace}
\newcommand{\pre}{Prec@$k$\xspace}

\newcommand{\std}{{\sc std}\xspace}
\newcommand{\h}{{\sc h}\xspace}
\newcommand{\ch}{{\sc ch}\xspace}
\newcommand{\iind}{{\sc i}\xspace}
\newcommand{\dunn}{{\sc d}\xspace}
\newcommand{\s}{{\sc s}\xspace}
\newcommand{\db}{{\sc db}\xspace}
\newcommand{\xb}{{\sc xb}\xspace}
\newcommand{\sd}{{\sc sd}\xspace}

\newcommand{\emmes}{{\sc EM}\xspace}
\newcommand{\mv}{{\sc MV}\xspace}

\newcommand{\ireos}{{\sc IREOS}\xspace}
\newcommand{\udr}{{\sc UDR}\xspace}
\newcommand{\mc}{{\sc MC}\xspace}
\newcommand{\mcs}{{\sc MC$_S$}\xspace}

\newcommand{\hits}{{\sc HITS}\xspace}
\newcommand{\ens}{{\sc Ens}\xspace}

\newcommand{\hitsc}{{\sc HITS-auth}\xspace}
\newcommand{\ensc}{{\sc Ens-pseudo}\xspace}
\newcommand{\hitscs}{{\sc HITS-au}\xspace}
\newcommand{\enscs}{{\sc Ens-pse}\xspace}
\newcommand{\rand}{{\sc Random}\xspace}
\newcommand{\ifor}{{\sc iForest-r}\xspace}

\newcommand{\rnd}{{\sc Rnd}\xspace}
\newcommand{\ifr}{{\sc iF}\xspace}
\newcommand{\best}{{\sc Best}\xspace}


\newcommand{\bX}{\bold{X}}
\newcommand{\bx}{\mathbf{x}}
\newcommand{\ba}{\mathbf{a}}
\newcommand{\bbeta}{\boldsymbol{\beta}}
\newcommand{\bmu}{\boldsymbol{\mu}}
\newcommand{\bchi}{\boldsymbol{\chi}}
\newcommand{\bPhi}{\boldsymbol{\Phi}}
\newcommand{\bphi}{\boldsymbol{\phi}}
\newcommand{\btheta}{\boldsymbol{\theta}}

\newcommand{\reminder}[1]{{\textsf{\textcolor{red}{[TODO: #1]}}}}
\newcommand{\bnote}[1]{{\textsf{\textcolor{blue}{[#1]}}}}
\newcommand{\summary}[1]{{\textsf{\textcolor{red}{#1}}}}

\newcommand{\hide}[1]{}


\newcommand{\Dtrn}{\mathcal{D}_\mathrm{trn}}
\newcommand{\Dval}{\mathcal{D}_\mathrm{val}}
\newcommand{\Dtest}{\mathcal{D}_\mathrm{test}}
\newcommand{\Daug}{\mathcal{D}_\mathrm{aug}}
\newcommand{\Ztrn}{\mathcal{Z}_\mathrm{trn}}
\newcommand{\Zaug}{\mathcal{Z}_\mathrm{aug}}
\newcommand{\Zval}{\mathcal{Z}_\mathrm{val}}
\newcommand{\ztrn}{\mathbf{z}_\mathrm{trn}}
\newcommand{\zval}{\mathbf{z}_\mathrm{val}}
\newcommand{\zaug}{\mathbf{z}_\mathrm{aug}}
\newcommand{\za}{\mathbf{z}_\mathrm{a}}
\newcommand{\xtrn}{\mathbf{x}_\mathrm{trn}}
\newcommand{\xtrntilde}{\mathbf{\tilde{x}}_\mathrm{trn}}
\newcommand{\xaug}{\mathbf{x}_\mathrm{aug}}
\newcommand{\xaugtilde}{\mathbf{\tilde{x}}_\mathrm{aug}}
\newcommand{\xval}{\mathbf{x}_\mathrm{val}}
\newcommand{\faug}{f_\mathrm{aug}}
\newcommand{\fdet}{f_\mathrm{det}}
\newcommand{\Ltrn}{\mathcal{L}_\mathrm{trn}}
\newcommand{\Lval}{\mathcal{L}_\mathrm{val}}
\newcommand{\Laug}{\mathcal{L}_\mathrm{aug}}

\author{Boje Deforce\thanks{KU Leuven, Belgium. (\texttt{boje.deforce@kuleuven.be})}
\and Meng-Chieh Lee\thanks{Carnegie Mellon University, USA.}
\and Bart Baesens\footnotemark[1]
\thanks{University of Southampton, United Kingdom.}
\and Estefanía Serral Asensio\footnotemark[1]
\and Jaemin Yoo\thanks{Korea Advanced Institute of Science and Technology, Korea.}
\and Leman Akoglu\footnotemark[2]
}
\title{End-To-End Self-Tuning Self-Supervised Time Series Anomaly Detection }


\date{}

\maketitle


\fancyfoot[R]{\scriptsize{Copyright \textcopyright\ 2025 by SIAM\\
Unauthorized reproduction of this article is prohibited}}





\begin{abstract}

Time series anomaly detection (TSAD) finds many applications such as monitoring environmental sensors, industry KPIs, patient biomarkers, etc. A two-fold challenge for TSAD is a versatile and unsupervised model that can detect various \textit{different types} of time series anomalies (spikes, discontinuities, 
trend shifts, etc.) \textit{without any labeled data}. Modern neural networks have outstanding ability in modeling complex time series. Self-supervised models in particular tackle unsupervised TSAD by transforming the input via various augmentations to create pseudo 
anomalies for training. However, their performance is sensitive to the choice of augmentation, which is hard to choose in practice, while
there exists no effort in the literature on data augmentation tuning for TSAD without labels.  Our work aims to fill this gap. We introduce \method 
for \textit{TSA ``on autoPilot''}, which can \textit{(self-)tune} augmentation hyperparameters end-to-end. It stands on two key components: a differentiable  augmentation architecture and an unsupervised validation loss to effectively assess the alignment between augmentation type and anomaly type.
Case studies show \method's ability to effectively select the (discrete) augmentation type and associated (continuous) hyperparameters.
 In turn, it outperforms established baselines, including SOTA self-supervised models, on diverse TSAD tasks exhibiting different anomaly types.\looseness=-1  
\end{abstract}


\section{Introduction}
\label{sec:intro}

Anomaly detection (AD) is a critical task in various domains such as cybersecurity, healthcare, and finance.
AD is especially important in time series to ensure system safety and reliability.
Thus, there exists a large body of work on time series AD as presented in various surveys~\cite{blazquez2021review,gupta2013outlier}. Recent progress on self-supervised learning (SSL) has changed the field of AD, offering significant improvements over traditional unsupervised (or one-class) learning approaches.
The main advantage of SSL lies in its ability to self-generate labeled samples, i.e., \textit{pseudo} anomalies, within the input space.
This enables a focused exploration of a plausible subspace based on the semantics reflected in the pseudo anomalies, rather than an exhaustive, impractical search of the entire space.
At the heart of SSL-based AD are data augmentation functions. 
These functions are used to create pseudo labels for the (self-)supervised training of an anomaly detector, such as by predicting whether the input is augmented or not \cite{Li21CutPaste}, which augmentation function is used \cite{Golan18GEOM}, or using contrastive learning \cite{Tack20CSI}.
In all these approaches, the success of SSL-based AD highly depends on the degree to which the augmented data mimics the true anomalies, as outlined in \cite{yoo2023data}.
There exist few approaches for tuning data augmentation functions without labels, however, they have limitations.
Some rely on non-differentiable validation losses \cite{Yoo23DSV}, which are unsuitable for end-to-end learning frameworks.
Others only address continuous hyperparameters, neglecting discrete ones, which are left for manual adjustment \cite{Yoo23End}.
For instance, in the image domain, the CutOut augmentation cannot mimic semantic class anomalies (e.g. cats vs. cars), 
no matter how one tunes its (continuous) hyperparameters \cite{Yoo23End}, mainly due to the mismatch in the discrete choice (i.e., augmentation type). Moreover, none of these efforts addresses anomaly detection for time series data, which is the focus of our work.

In this paper we introduce \method, a novel approach for SSL-based time series anomaly detection (\underline{TSA}D) ``on auto\underline{P}ilot'' equipped with end-to-end hyperparameter tuning. Effectively tuning both discrete and continuous hyperparameters of augmentation enables \method to be an automated anomaly detector that is most suitable for a given task. \method stands on two main components: ($i$) a differentiable parameterized augmentation model and ($ii$)  an unsupervised validation loss that measures the alignment between the augmented data and the unlabeled test data, quantifying the extent to which the former mimics the true anomalies.

\noindent We summarize our main contributions as follows:
\begin{compactenum}
    \item \textbf{Problem:} Our work is the first attempt to tune both~discrete and continuous hyperparameters of data augmentation in SSL-based TSAD, without labels at training time.
    \item \textbf{New TSAD Method:} We propose \method\footnote{All code and datasets used in this work are available at the following repository: {\url{https://tinyurl.com/hhfdrrtk}}.}
    , which accommodates 
    various time series anomaly types
    and enables automatic tuning of related hyperparameters (magnitude, duration, \ldots) with a differentiable validation loss quantifying alignment of augmented and unlabeled test data. 
    \item \textbf{Effectiveness:} By carefully selecting augmentation type and its hyperparameters, our self-tuning \method outperforms existing unsupervised and self-supervised approaches, 
    including the SOTA NeuTraL-AD \cite{qiu2021neutral} which also employs learnable augmentations, across diverse TSAD tasks.
\end{compactenum}

\begin{figure*}[!ht]
    \centering
    \hspace*{-0.21in}
    \includegraphics{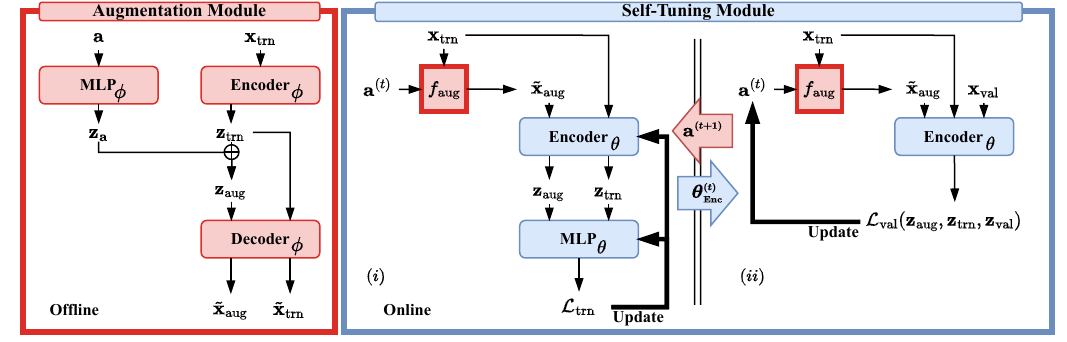}
   \vspace{-0.2in}
    \caption{Our \method framework for end-to-end self-tuning TSAD. \textbf{Left:} Offline trained, differentiable augmentation model $\faug(\cdot;\bphi)$ takes as input the normal data and augmentation hyperparameter(s) $\ba$, and outputs pseudo-anomalies $\xaugtilde$. \textbf{Right:} Self-tuning engine incorporates the pre-trained $\faug$ (with parameters $\bphi$ frozen), alternating between two phases: ($i$) detection phase --  given $\ba^{(t)}$ at iteration $t$, estimate parameters $\btheta^{(t)}$ of detector $\fdet$ (consisting of Encoder$_\theta$ and discriminator MLP$_\theta$), by optimizing $\Ltrn$ (classification loss); ($ii$) alignment phase -- given $\fdet^\mathrm{enc}(\cdot;\btheta^{(t)})$, update augmentation (governed by $\ba$) to better align the embedded time series $\ztrn \cup \zaug$ with $\zval$ in the learned discriminative space. Note that $\xval$ contains both normal and anomalous time series, but labels are \textit{not} known or used at any point during training time.}
    \label{fig:tsad-framework}
    \vspace{-0.1in}
\end{figure*}

\section{Preliminaries}
\label{sec:prelim}

\subsection{Time Series Anomaly Detection}

This work focuses on \emph{time series anomaly detection}.
Consider a univariate time series $\bx = \{x_1, x_2, \ldots, x_K\}$, where $\bx$ is a sequentially ordered collection of $K$ data points. Each  $x_k \in \mathbb{R}$ 
corresponds to a scalar observation at each time step. 
Then, let $\Dtrn$ be a set of training data containing only normal time series, and $\Dtest$ be a set of unlabeled test data containing both normal and anomalous time series.
Given $\Dtrn$ and $\Dtest$, the problem is to detect all anomalous time series in $\Dtest$, i.e., to assign the correct label $y \in \{ -1, +1 \}$ for $\bx \in \Dtest$.

\subsection{Self-supervised Anomaly Detectors}

Let $\faug(\cdot;\bphi)$ be a data augmentation function, such as rotation or cut-out in the image domain (a discrete choice), that outputs \textit{pseudo} anomalies. A given $\faug$ also exhibits continuous hyperparameter(s), such as angle (for rotation) or width and height (for cutout). Then, a popular approach \cite{Li21CutPaste} is to train a detector model $\fdet(\cdot;\btheta)$ to classify between normal data $\Dtrn$ and augmented data $\Daug = \{ \faug(\bx) \mid \bx \in \Dtrn \}$.
The working assumption is for $\fdet$ to predict the (unknown) anomalies in test data as \emph{augmented}. As such, performance relies on how well the choice of augmentation type (discrete) as well as the choice(s) of its (continuous) hyperparameter(s)  mimic the anomalies in $\Dtest$.
There are various other ways of using SSL for anomaly detection, based on the specific objective function and how the pseudo labels are generated \cite{Golan18GEOM,Tack20CSI}, which are similar to each other in principle.

\subsection{Data Augmentation on Time Series}

To apply SSL to time series data, we need an augmentation function specifically designed for time series. For a given time series $\bx \in \Dtrn$, $\faug(\bx; \ba)$ transforms $\bx$ based on hyperparameters $\ba \in \mathcal{A}^P$, $P \geq 1$, where $\mathcal{A}^P$ represents the domain of possible hyperparameter values.
For time series data, augmentations could involve mean shifts, trend injections, or other transformations relevant to the time-series domain, with $\mathcal{A}^P$ enclosing the respective hyperparameter space for a given transformation. We describe augmentation in more detail in Sec. \ref{sec:faug}.

\vspace{-0.075in}
\subsection{Wasserstein Distance}

The Wasserstein distance \cite{ArjovskyCB17}, a distance measure between probability distributions, of order $p$ between any two marginal distributions $\mu$ and $\nu$ is given as
\begin{equation}\label{eq:wass}
    W_p(\mu, \nu) = \biggl( \inf_{\gamma \in \Gamma(\mu, \nu)} \mathbb{E}_{(x, y) \sim \gamma} [ d(x, y)^p ] \biggr)^{1 / p},
\end{equation}
where $\Gamma(\mu, \nu)$ is the set of all joint distributions (or couplings) with marginals  $\mu$ and $\nu$, respectively.
That is, $\gamma$ satisfies 
two conditions:
$
    \int \gamma(x, y)dy = \mu(x)$ and $\int \gamma(x, y)dx = \nu(y)
    $.
%
However, computing $W_p$ directly is often computationally challenging. Thus, we employ the Sinkhorn algorithm to feasibly apply the Wasserstein distance in the machine learning context, which provides an efficient approach to approximate it via entropy regularization \cite{Cuturi13Sinkhorn}.


\vspace{-0.075in}
\section{\method: Time Series Anomalies on AutoPilot}
\label{sec:measures}

There are two notable challenges that need to be addressed for automatic selection of both discrete and continuous hyperparameters for SSL-based TSAD:
\vspace{-0.1in}
\subsection*{Challenges}

\begin{compactenum}
    \item[\textbf{C1.}] \textbf{Differentiable Augmentation:} Developing an augmentation function that is differentiable with respect to its hyperparameters, enabling gradient-based optimization.
    \item[\textbf{C2.}] \textbf{Comparable Validation Loss:} Formulating a validation loss that effectively quantifies alignment between $\Dtrn \cup \Daug$ and $\Dtest$ while being comparable \emph{across} different hyperparameter initializations.
\end{compactenum}

{
\setlength{\parindent}{0pt}
Our framework tackles \textbf{C1} and \textbf{C2} with two key ideas:
}

\vspace{-0.1in}
\subsection*{Main ideas}

\begin{compactenum}
    \item[\textbf{I1.}] \textbf{Differentiable Augmentation Module:} \method implements the augmentation function as an Encoder-Decoder neural network, $\faug$ parameterized by $\bphi$, capable of approximating the anomaly-generating mechanism conditioned on $\ba \in \mathcal{A}^P$. Importantly, this module is pre-trained independently and prior to the initiation of \textbf{I2}, establishing it as an \textit{offline} component of the framework.
    \item[\textbf{I2.}] \textbf{Self-Tuning Module:} At test time \textit{online}, \method iteratively refines the detector $\fdet$'s parameters $\btheta$ as well as augmentation hyperparameters $\ba$, through alternating detection and alignment phases. Alignment is performed on part of the unlabeled $\Dtest$, referred to as $\Dval$.
\end{compactenum}

{
\setlength{\parindent}{0pt}
Based on \textbf{I1} and \textbf{I2}, we propose \method, a self-tuning self-supervised TSAD framework demonstrated in Fig.~\ref{fig:tsad-framework}. 
}

\subsection{Differentiable Augmentation Module}\label{sec:faug}

\begin{figure}[!t]
    \centering
    \resizebox{\columnwidth}{!}{
    \includegraphics{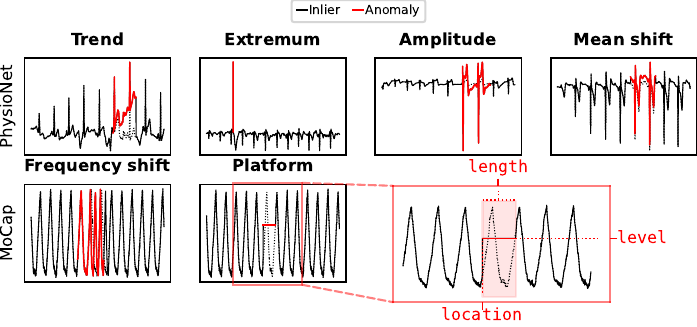}
    }
    \vspace{-0.3in}
    \caption{Examples of six different types of time series anomalies; (black) original real-world time series, (red) pseudo anomalies generated by $g$. 
    }
    \label{fig:faug}
\end{figure}

\paragraph{Anomaly Injection Scheme:~}
We accommodate six types of anomalies that are common in real-world time series; namely, trend, extremum, amplitude, mean shift, frequency shift, and platform. 
A detailed description of the anomalies is provided in Appx.~\ref{ssec:tsatypes} and for brevity, we illustrate them by examples in Fig.~\ref{fig:faug} with red lines.
Each anomaly type has three hyperparameters; including its starting position ({\tt location}), duration ({\tt length}), and severity ({\tt level}) as shown in Fig.~\ref{fig:faug}. 
Extremum captures a spike and only has two hyperparameters as its duration is always $1$. 
Based on $\Dtrn$, the anomaly generation scheme $g$ creates an augmented dataset $\Daug = \{ g(\xtrn; \ba) \mid \xtrn \in \Dtrn, \ba \sim \mathcal{A}^P \}$, where $\ba$ is the vector of augmentation hyperparameters, uniformly randomly sampled from the domain $\mathcal{A}^P$.

\paragraph{Model Design:~}
To build an augmentation model $\faug$ for time series, we use a convolutional neural network (CNN) to encode the 
input time series $\xtrn$ into the feature map $\ztrn$.
We then encode the augmentation hyperparameters $\ba$ into $\za$, which has the same shape as $\ztrn$, with a multilayer perceptron (MLP).
Since the feature map $\ztrn$ generated by the CNN encoder keeps the positional information of the original time series, adding $\za$ to $\ztrn$ ensures that only the part with the desired {\tt location} and {\tt length} in $\zaug$ is manipulated.
To ensure that the feature maps $\zaug$ and $\ztrn$ are in the same embedding space, they share the same decoder to reconstruct back to the time series $\xaug=g(\xtrn;\ba)$ and $\xtrn$, respectively.
As such, the loss function of $\faug$ is based on the reconstruction of both $\Dtrn$ and $\Daug$:

\vspace{-0.2in}
{\small
\begin{align}
    \Laug = 
    \sum_{\substack{\xtrn \in \Dtrn \\ \ba \sim \mathcal{A}^P}}
    \hspace{-0.1in}
     \big(
     {(\xtrn - \xtrntilde)}^{2}
     + 
{(g(\xtrn,\ba) - \xaugtilde)}^{2} \big)
\end{align}
}
where $\xaugtilde$ denotes the output of $\faug(\xtrn;\ba)$ for the hyperparameters sampled by $g$ (see Fig.~\ref{fig:tsad-framework} left).

\begin{algorithm}[!t]
\caption{Self-Tuning Module of \method}
{\small{
\begin{algorithmic}[1]
\Statex \textbf{Input:} Train
\& val. data $\Dtrn$\& $\Dval$, pre-trained aug. model $\faug(\cdot;\bphi)$, detector $\fdet(\cdot;\btheta)$, max epoch $T$, \#inner-loops $L$, and step sizes $\alpha$ and $\beta$ 
\Statex \textbf{Output:} Optimized aug. hyperparameters $\ba^*$ and optimized parameters $\btheta^*$ corresponding to the lowest $\Lval^*$
\State $\ba^{(0)} \sim \mathrm{Uniform}(\mathcal{A}^P)$ 
\For {$t = 0, 1, \ldots, T-1$}
    \State $\btheta' \gets \btheta - \alpha \nabla_{\btheta}\Ltrn(\btheta, \ba^{(t)})$ \textbf{for} $l$ \textbf{in} $L$ \textcolor{blue}{\Comment{Phase ($i$)}}
    \State $\Ztrn \gets \{\fdet^\mathrm{enc}(\bx) \mid \bx \in \Dtrn\}$ \textcolor{blue}{\Comment{Phase ($ii$) starts}}
    \State $\Zaug \gets \{\fdet^\mathrm{enc}(\faug(\bx, \ba^{(t)})) \mid \bx \in \Dtrn\}$ 
    \State $\Zval \gets \{\fdet^\mathrm{enc}(\bx) \mid \bx \in \Dval\}$ 
    \State $\Ztrn, \Zaug, \Zval \gets \mathrm{norm.}(\Ztrn, \Zaug, \Zval)$ 
    \State $\ba^{(t+1)} \gets \ba^{(t)} - \beta \nabla_{\ba}\Lval(\Ztrn, \Zaug, \Zval)$ 
    \State $\btheta \gets \btheta'$ \textcolor{blue}{\Comment{Phase ($ii$) ends}}
    \State \textbf{if} {$\Lval^{(t)} < \Lval^*$} \textbf{then } 
        $\ba^* \gets \ba^{(t+1)}$; $\;$
$\btheta^* \gets \btheta$ \textbf{ end if} 
\EndFor
\State Return $\ba^*$ and $\btheta^*$
\end{algorithmic}\label{alg:self-tuning}
}}
\end{algorithm}
\setlength{\textfloatsep}{0.1in}

\subsection{Self-Tuning Module}

Central to \method is the self-tuning module, which operates by iteratively refining the detector's parameters, $\btheta$, and the augmentation hyperparameters, $\ba$. The process is structured into two phases: detection and alignment (see Fig.~\ref{fig:tsad-framework} right).
The overall algorithm is given in Algo.~\ref{alg:self-tuning}.

\paragraph{Phase ($i$) Detection:~} 

This phase focuses on estimating 
the parameters $\btheta^{(t)}$ of the detector $\fdet$ (comprising of an encoder $\fdet^\mathrm{enc}$ and a discriminator $\fdet^\mathrm{mlp}$)
by minimizing the cross-entropy loss $\Ltrn$.
This aims to classify between the normal samples $\xtrn$ and the augmented pseudo anomalies $\xaugtilde$ by their embeddings $\Ztrn$ and $\Zaug$, where 
$\Ztrn = \{ \fdet^\mathrm{enc}(\bx) \mid \bx \in \Dtrn \}$ and $\Zaug = \{\fdet^\mathrm{enc}(\faug(\bx; \ba^{(t)})) \mid \bx \in \Dtrn\}$ denote the embeddings of the training data and augmented data, respectively, given the current \textit{fixed} $\ba^{(t)}$ at iteration $t$. 
Note that the parameters $\bphi$ of the augmentation model $\faug$ are frozen throughout this phase.

\paragraph{Phase ($ii$) Alignment:~} 

Subsequently, the alignment phase adjusts $\ba$ to optimize the unsupervised differentiable validation loss $\Lval$, computed based on the embeddings from the now-updated $\fdet^\mathrm{enc}$.  $\Lval$'s objective is to measure the degree of alignment between $\Dtrn \cup \Daug$ and $\Dval$ in the embedding space, 
as expressed by the Wasserstein distance 
in \eqref{eq:wass}. This metric is chosen for its effectiveness in capturing the overall distributional discrepancies between datasets, offering a more complete comparison than mere point-wise metrics \cite{ArjovskyCB17}, which is especially important in TSAD given the often subtle nature of time series anomalies. The embeddings are normalized to ensure scale invariance before being passed to $\Lval$, so as to avoid the trivial solution of achieving alignment by setting all embeddings to the zero vector \cite{Yoo23End}. Lastly, as the embeddings $\{\Ztrn, \Zaug, \Zval\}$ in $\Lval$ are obtained through the updated $\fdet^{\mathrm{enc}}$, the optimization process needs to track the change in $\btheta$ caused by the update of $\ba$. As such, \method uses a second-order optimization process, similar to \cite{Yoo23End}.


\paragraph{Augmentation Type Selection:~}

While Algo.~\ref{alg:self-tuning} describes continuous hyperparameter tuning, the discrete hyperparameter (i.e. augmentation type) selection, corresponding to a specific anomaly type, is done through grid search as the number of anomaly types is finite. Hence, we initialize \method for different augmentation types and compare $\Lval$ across types to select the one that yields the best alignment. The idea is that the wrong augmentation type will have poor alignment, while one that reflects the true anomalies in $\Dtest$ will result in better alignment, granted proper tuning of the continuous hyperparameters through Algo. \ref{alg:self-tuning}.

\vspace{-0.075in}
\section{Experiments}
\vspace{-0.05in}
\label{sec:experiment}

In this section, we aim to answer the following questions:
\begin{compactenum}[{Q}1.]
    \item \textbf{Quantitative Comparison}: How well does \method perform on TSAD against established baselines?
    \item \textbf{Qualitative Analysis}: How well does \method tune the augmentation hyperparameters?
    \item \textbf{Ablation Studies}: Which design decisions in \method drive its effectiveness?
\end{compactenum}

\vspace{-0.1in}
\subsection{Experimental Setup}\label{sec:exp_setup}

\paragraph{Datasets:~} We evaluate \method on six distinct TSAD tasks. Four of these are conducted in a \emph{controlled} environment, while the remaining two are \emph{natural}. In the controlled setting, the anomaly types are manually injected in $\Dtest$ based on the types discussed in Sec.~\ref{sec:faug}. The controlled environment allows for a thorough assessment of \method's ability to perform continuous hyperparameter tuning. For the natural environment, the anomaly types in $\Dtest$ are a priori unknown and it is the goal of \method to find the type that yields best alignment between $\Dtrn \cup \Daug$ and $\Dval$ (part of $\Dtest$), expressed by $\Lval$. Further details are provided in Appx.~\ref{ssec:tsatypes}~and~\ref{ssec:datasets}.


For controlled tasks, we use the 2017 PhysioNet Challenge dataset \cite{Clifford2017physionet}, which comprises real-world ECG recordings.
Table \ref{tab:datasets} shows a selection of four different controlled TSAD tasks, constructed by manually injecting different anomaly profiles into the PhysioNet dataset. 
For PhysioNet A and B, given the anomaly type (Platform), the task is to infer or tune, respectively, the hyperparameter(s) {\tt level} only and both {\tt level} and {\tt length}, while anomaly location is random (hence not tuned). For PhysioNet C and D, the respective tuning tasks are the same but for a different anomaly type (Trend).
We choose Platform and Trend anomalies to represent, respectively, range-bound and range-exceeding anomalies. Additional variations of PhysioNet with other injected anomaly types, such as Mean shift and Extremum, are discussed in Appx.~\ref{ssec:datasets}.

The natural TSAD tasks are derived from the CMU Motion Capture (MoCap) dataset
\footnote{http://mocap.cs.cmu.edu/}.
We consider the walking signal as normal data, and the signals of jumping and running as anomalies.
To generate normal signals, we stitch the walking signals by identifying the start and end points of each gait phase and add random noise; whereas to generate anomalous ones, we stitch walking or running signals at a random location in the normal signal.
This yields two distinct TSAD tasks as shown in Table \ref{tab:datasets}.
Different from PhysioNet A--D where we only tune the continuous hyperparameter(s) for the given (discrete) anomaly type, here for MoCap A and B, we aim to tune {\em both} the unknown anomaly type that corresponds to Jump and Run behavior, respectively, as well as the (continuous) hyperparameter {\tt level} while {\tt location} and {\tt length} take random values.\looseness=-1


\begin{table}[!t]
\caption{Anomaly profile of different TSAD tasks.}
\centering
\resizebox{0.9\linewidth}{!}{%
\begin{tabular}{l | c |cccc}
    \toprule
      \multicolumn{2}{l}{Dataset}   & {\tt Type} & {\tt Level} & {\tt Location} & {\tt Length} \\
      \midrule
      \multirow{4}{*}{\rotatebox[origin=c]{90}{\small PhysioNet}} & PhysioNet A  & Platform & Fixed &  Random & Random \\
      & PhysioNet B  & Platform & Fixed &  Random & Fixed \\
      & PhysioNet C  & Trend & Fixed &  Random & Random \\
      & PhysioNet D  & Trend & Fixed &  Random & Fixed \\
      \midrule
      \multirow{2}{*}{\rotatebox[origin=c]{90}{\fontsize{6.5}{10}\selectfont MoCap}} & MoCap A  & Jump & Fixed &  Random & Random \\
      & MoCap B  & Run & Fixed &  Random & Random \\
     \bottomrule
\end{tabular}
}
\label{tab:datasets}
\end{table}


\begin{table*}[!t]
 \caption{Detection peformance of baselines w.r.t. $F_1$ and AUROC on test data across six TSAD tasks. Four tasks are performed in a controlled setting (manually injected anomaly type, cf.~Sec.~\ref{sec:faug}), based on the PhysioNet ECG data. Remaining two TSAD tasks exhibit natural anomalies (unknown real-world anomaly type), based on the MoCap data. \textbf{\method outperforms most baselines and has the best (lowest) average rank w.r.t. both $F_1$ and AUROC, with low standard deviation (in parentheses).}
 }
    \centering
    \resizebox{\textwidth}{!}{%
    \begin{tabular}{l | cc c cc c cc c cc c cc c cc | cc}
        \toprule
        \multirow{2}{*}{Methods} 
              & \multicolumn{2}{c}{PhysioNet A} 
            & & \multicolumn{2}{c}{PhysioNet B} 
            & & \multicolumn{2}{c}{PhysioNet C} 
            & & \multicolumn{2}{c}{PhysioNet D} 
            & & \multicolumn{2}{c}{MoCap A}
            & & \multicolumn{2}{c}{MoCap B} \vline
            & \multicolumn{2}{c}{Avg. Rank}
            \\
        \cmidrule{2-3} \cmidrule{5-6} \cmidrule{8-9} \cmidrule{11-12} \cmidrule{14-15} \cmidrule{17-18} \cmidrule{19-20}
        & $F_1$ & AUROC &  & $F_1$ & AUROC & & $F_1$ & AUROC & & $F_1$ & AUROC & & $F_1$ & AUROC &  & $F_1$ & AUROC & $F_1$ & AUROC \\
        \midrule
        OC-SVM          & 0.182 & 0.468 & & 
                          0.182 & 0.472 & & 
                          0.373 & 0.803 & & 
                          0.393 & 0.806 & & 
                          \textbf{1.000} & \textbf{1.000} & & 
                          0.546 & 0.806 &
                          7.2 (3.7) & 6.8 (3.3) \\
        LOF             & \underline{0.999} & \textbf{1.000} & & 
                          \underline{0.999} & \textbf{1.000} & & 
                          0.354 & 0.738 & & 
                          0.358 & 0.725 & & 
                          0.196 & 0.506 & & 
                          0.221 & 0.577 &
                          6.8 (3.9) & 6.5 (4.4) \\
        ARIMA           & 0.885 & 0.960 &&
                          0.829 & 0.965 &&
                          \textbf{0.991} & \textbf{0.999} &&
                          \textbf{0.999} & \textbf{0.999} &&
                          0.870 & 0.955 &&
                          0.225 & 0.537 &
                          4.3 (3.2) & 4.7 (3.7) \\
        IF              & 0.255 & 0.587 & & 
                          0.232 & 0.576 & & 
                          0.183 & 0.402 & & 
                          0.182 & 0.356 & & 
                          0.864 & 0.965 & & 
                          0.342 & 0.758 &
                          8.8 (1.7) & 8.7 (1.9) \\
        MP              & 0.812 & 0.743 & & 
                          0.812 & 0.744 & & 
                          0.280 & 0.712 & & 
                          0.284 & 0.734 & & 
                          \textbf{1.000} & \textbf{1.000} & & 
                          \textbf{1.000} & \textbf{1.000} &
                          5.0 (3.6) & 4.8 (3.4) \\
        \midrule
        EncDec-LSTM     & 0.190 & 0.508 & & 
                          0.190 & 0.508 & & 
                          0.415 & 0.812 & & 
                          0.442 & 0.819 & & 
                          \underline{0.980} & \underline{0.999} & & 
                          \underline{0.909} & \underline{0.996} &
                          6.5 (2.0) & 6.7 (1.9) \\
        SR-CNN          & 0.965 & \underline{0.990} & & 
                          0.964 & \underline{0.998} & & 
                          \underline{0.983} & \textbf{0.999} & & 
                          \underline{0.991} & \textbf{0.999} & & 
                          0.302 & 0.700 & & 
                          0.214 & 0.512 &
                          5.2 (4.2) & 4.8 (4.5) \\
        USAD            & 0.183 & 0.425 & & 
                          0.184 & 0.428 & & 
                          0.430 & 0.822 & & 
                          0.409 & 0.828 & & 
                          \textbf{1.000} & \textbf{1.000} & & 
                          \textbf{1.000} & \textbf{1.000} &
                          5.5 (4.0) & 5.7 (4.5) \\
        NeuTraL-AD      & 0.211 & 0.732 & & 
                          0.263 & 0.679 & & 
                          0.561 & 0.868 & & 
                          0.526 &  0.862 & &  
                          \textbf{1.000} & \textbf{1.000} & & 
                          \textbf{1.000} & \textbf{1.000} &
                          \underline{4.2} (2.9) & \underline{3.8} (2.5) \\
        TimeGPT         & 0.327 & 0.714 & & 
                          0.318 & 0.711 & &
                          0.218 & 0.580 & & 
                          0.217 & 0.525 & &
                          0.348 & 0.743 & & 
                          0.385 & 0.683 &
                          8.0 (1.9) & 8.3 (1.6) \\
        \midrule
        \method (ours)  & \textbf{1.000} & \textbf{1.000} & & 
                          \textbf{1.000} & \textbf{1.000} & & 
                          0.973 & \textbf{0.999} & & 
                          \underline{0.991} & \underline{0.998} & & 
                          0.889 & 0.969 & & 
                          \textbf{1.000} & \textbf{1.000} &
                          \textbf{2.3} (2.0) & \textbf{2.2} (2.0) \\
        \bottomrule
    \end{tabular}
    }
   \label{tab:results}  
\end{table*}

\paragraph{Baselines:~}

We compare \method with a selection of established baselines, including traditional and deep learning-based methods with demonstrated efficacy in TSAD \cite{schmidl22Survey}. The traditional methods consist of different modeling approaches; namely, One-Class Support Vector Machines (\textbf{OC-SVM}) 
\cite{Scholkopf99SVM}; 
Local Outlier Factor (\textbf{LOF})
\cite{Breunig00LOF};
(\textbf{ARIMA}) \cite{box2015time};
Isolation Forest (\textbf{IF}) 
\cite{Liu08IF};
and the Matrix Profile (\textbf{MP})
\cite{yeh2016mp}.
On the deep learning side, we benchmark against the Encoder-Decoder LSTM (\textbf{EncDec-LSTM}) 
\cite{malhotra2016lstm}; the Spectral Residual Convolutional Neural Network (\textbf{SR-CNN}) 
\cite{Hansheng19SRCNN}; the Unsupervised Anomaly Detection (\textbf{USAD}) for TSAD 
\cite{Audibert20USAD};
and a recent time series foundation model (\textbf{TimeGPT}) \cite{garza2023timegpt}.
Lastly, we include a state-of-the-art competing method which learns augmentations in the embedding space, called Neural Transformation Learning for (TS)AD (\textbf{NeuTraL-AD}) \cite{qiu2021neutral}.
This diverse set of baselines allows for a comprehensive analysis across different approaches within the TSAD domain.


\paragraph{Model Configurations:~}

The Encoder$_\phi$ in $\faug$ and Encoder$_\theta$ in $\fdet$ are constructed using 1D CNN blocks \cite{bai2018cnn} (transposed 1D CNN for Decoder$_\phi$) for efficient temporal feature extraction. We choose the number of epochs $T$ to allow sufficient time for the convergence of $\ba$, with empirical evidence suggesting that $T = 100$ typically suffices. For the number of inner-loops $L$, we set $L=5$, aligned with \cite{Yoo23End}, such that $\fdet$ has adequate time to learn effective discriminative embeddings for $\Daug$ and $\Dtrn$. An overview of the configuration for \method and all baselines is provided in Appx.~\ref{ssec:config}.\looseness=-1

\paragraph{Evaluation Metrics:~}
Our method calculates anomaly scores on an entire sequence level $\mathbf{x}$, similar to \cite{qiu2021neutral}. This is a different set-up compared to novelty detection in time series which typically operates on a point level. Detection on a sequence level can be especially important to spot long-term patterns (e.g. Trend anomalies). As such, we use the $F_1$ score and the Area Under the Receiver Operating Characteristic Curve (AUROC) as key performance metrics to quantify detection capability of anomalous sequences. All results are reported on the unseen $\Dtest$. 
%
We determine the optimal $F_1$ score, by enumerating all possible thresholds, given by the anomaly scores for a given segment $\mathbf{x}$. We then compute the corresponding precision and recall for each threshold and select those that yield the highest $F_1$. As such, AUROC provides a balanced view whereas $F_1$ shows the optimal scenario. 
Both metrics range from 0 to 1, with higher values indicating superior performance.\looseness=-1

\subsection{Quantitative Results}

Table \ref{tab:results} provides the detection results for all six TSAD tasks 
described in Sec.~\ref{sec:exp_setup}. 
{\textbf{\method ranks the best overall}} in terms of both $F_1$ and AUROC. This shows that detector $\fdet(\cdot;\btheta^*)$, trained through the alternating mechanism of \method, is able to generalize to unseen and unlabeled anomalies in $\Dtest$. 
{\textbf{While some competing methods perform strongly on a subset of tasks, they lack consistency across all TSAD tasks.
}} Among the traditional baselines, which perform subpar as compared to deep TSAD approaches, LOF performs very well in PhysioNet A and B, relatively poorly on C and D, but exhibits near-random performance on MoCap. This discrepancy can be attributed to the nature of the anomalies in the PhysioNet data, which are manually injected and often characterized by pronounced and abrupt changes as shown in Fig.~\ref{fig:faug}. These abrupt changes lead to large differences in local density, which makes LOF thrive. In contrast, the MoCap dataset contains anomalies that emerge more subtly, as one gradually transitions from walking to running or jumping, reflecting more natural variations in the data which, in turn, lead to more subtle differences in local density.
Similarly, SR-CNN and ARIMA perform strongly on the PhysioNet TSAD tasks, yet their effectiveness diminishes on MoCap tasks. Both methods exploit the abrupt changes present in the PhysioNet data but fail to detect the more subtle anomalies in MoCap B. This suggests that, while these methods work well for certain types of anomalies, their utility may be limited in scenarios where anomalies are more subtle.
%
On the contrary, reconstruction-based baselines such as EncDec-LSTM and USAD face challenges on PhysioNet due to its high variability among inliers, which complicates the task of accurate data reconstruction. Yet, these methods excel with the MoCap dataset, where its consistent near-periodic pattern (Fig.~\ref{fig:faug}, bottom) lends itself to more reliable reconstruction, thereby enhancing anomaly detection performance. Similarly, MP struggles with the noisy PhysioNet data but excels at discord discovery in MoCap. Finally, NeuTraL-AD, a state-of-the-art augmentation-based baseline, demonstrates proficiency in properly augmenting the inlier data within MoCap. However, its performance on the PhysioNet variations reveals some weaknesses. NeuTraL-AD struggles with the high variability in the PhysioNet inliers, inherent to real-world ECG signals. This suggests that the augmentation functions considered in NeuTraL-AD lack robustness when inliers are inherently noisy. We remark that only \method provides robust and consistent performance across all TSAD tasks, showcasing the effectiveness and generalizability of our proposed method.

\subsection{Qualitative Results}

\begin{figure*}[!th]
    \centering
    \includegraphics{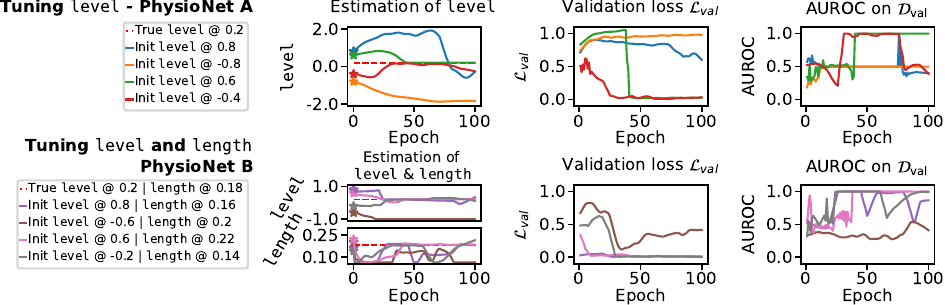}
    \vspace{-0.1in}
    \caption{\textbf{Tuning continuous augmentation hyperparameter(s)  
 with \method}. \textbf{Top:} Given Platform anomalies at true {\tt level} (red dashed line), various initializations converge accurately near the true value (left), following the minimized values of val. loss (center), and leading to high detection performance (right). \textbf{Bottom:} Multiple continuous hyperparameters, here both {\tt level} and {\tt length}, are accurately tuned to near true values (left), as guided by minimizing the val. loss (center), achieving high AUROC (right). Notice that the diverged results in both cases associate with high (or hard-to-optimize) val. loss, which help us  effectively reject low performance models.\looseness=-1}
    \label{fig:platform-estimation}
    \vspace{-0.1in}
\end{figure*}

A key contributor to \method's consistent performance is the validation loss through which \method automatically learns the augmentation hyperparameters $\ba$. Once $\ba$ is determined, the task essentially reduces to a supervised learning problem. Next, we show that \method not only {\textbf{effectively tunes the continuous augmentation hyperparameters $\ba$}}, but also that its validation loss guides the \textbf{{accurate selection of the discrete hyperparameter (i.e. anomaly type)}}. 

\paragraph{Controlled Environment:~}

Consider PhysioNet A, where we aim to tune the continuous hyperparameter $a$, i.e. {\tt level}, of the Platform anomalies present in $\Dtest$. That is, the {\tt level} in $\Dtest$ is fixed and tuning aims to estimate its value using \method while the other hyperparameters ({\tt location, length}) are randomized. Fig.~\ref{fig:platform-estimation} (top) shows \method's estimation process 
for different initializations of $a$. We observe that the initialization for $a \in \{-0.4, 0.6\}$ leads to the true $a$$=$$0.2$ (left). Simultaneously, the validation loss drops substantially once \method has arrived at the true $a$ (center). This is also reflected in the performance of $\fdet$ on $\Dval$ which soars upon estimation of the true $a$ (right). Conversely, the initialization for $a \in \{0.8, -0.8\}$ leads to a high validation loss, indicating poor alignment between  $\Dtrn \cup \Daug$ and  $\Dval$. Indeed, the performance of $\fdet$ on $\Dval$ now suffers from poor alignment. 

For PhysioNet B, we estimate both {\tt level} and {\tt length} while {\tt location} is randomized. Fig.~\ref{fig:platform-estimation} (bottom) demonstrates \method's ability to accurately estimate the {\tt level} and {\tt length}. While this is a by-product of our method, there are several real-world use-cases that can directly benefit from accurately learning the anomaly profile, from industrial equipment monitoring, to network security, and healthcare monitoring. 
This demonstrates \method's versatility given its capability of estimating continuous hyperparameters. See Appx.~\ref{ssec:addresults} for additional qualitative results on other anomaly types such as Trend, Mean shift, and Spike.

\begin{figure}[!t]
    \centering
    \resizebox{\columnwidth}{!}{
    \includegraphics{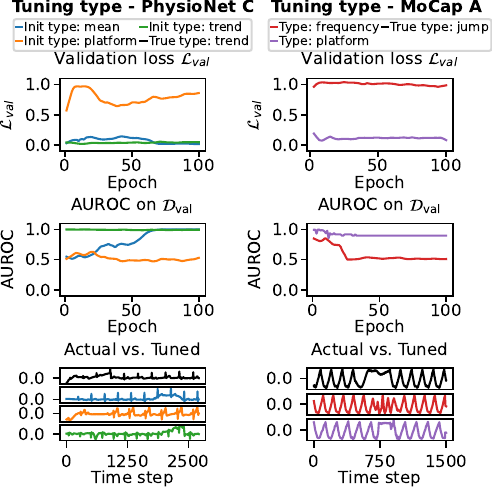}
    }
    \caption{\textbf{Tuning the discrete hyperparameter (anomaly type) with \method}. \textbf{Left:} Given true Trend anomaly (black), val. loss favors  both Trend (green) and Mean shift (blue) type, both with high AUROC (center) and high resemblance (right), and effectively rejects type Platform (orange) with poor performance.
    \textbf{Right:} For Jump anomalies in MoCap A  with unknown type (black), val. loss favors type Platform (purple) that leads to high AUROC (center) and mimics well the true anomaly (right), and effectively rejects type Frequency (red) with poor performance. 
    }
    \label{fig:mismatch}
\end{figure}

Further, Fig.~\ref{fig:mismatch} (left) showcases \method's ability to perform discrete hyperparameter selection. Here, \method has been initialized and trained with three different anomaly types (Mean shift, Platform, Trend) on PhysioNet C (true anomaly type is Trend). The validation loss clearly showcases a misalignment between the Platform and Trend types (top), also reflected in the AUROC of $\fdet$ on $\Dval$ (center). Note how the Mean shift anomaly type has a low $\Lval$ towards the end of the training epochs, which is also reflected in the AUROC on $\Dval$. Indeed, comparing \method tuned for Mean shift anomalies and for Trend anomalies show strong resemblances with the true underlying anomaly type (bottom). This shows that the true underlying anomaly type is not necessarily the only type that yields high alignment, and in turn a high-performing detector.
\paragraph{Natural Environment:~}

In MoCap datasets, the anomaly types are a priori unknown. As such, we initialize \method with different augmentation types (Frequency and Platform) to perform discrete hyperparameter selection. Fig.~\ref{fig:mismatch} (right) highlights its effectiveness as the validation loss clearly prefers one type over the other. Indeed, the natural anomalies defined by jumping signals in {MoCap A} have close resemblance to platform anomalies. 
{Discrete} hyperparameter optimization for MoCap B is similar, as given in Appx. \ref{ssec:addresults}.

\vspace{-0.1in}
\subsection{Ablation Studies}
\label{ssec:abl}
We showcase two ablation studies on the controlled PhysioNet data to support various design strategies of \method. Additional ablation studies on embedding normalization and second-order optimization are provided in Appx.~\ref{ssec:addresults}.

\paragraph{Validation Loss:~} 

\begin{figure}[!t]
    \centering
    \resizebox{\columnwidth}{!}{%
    \includegraphics{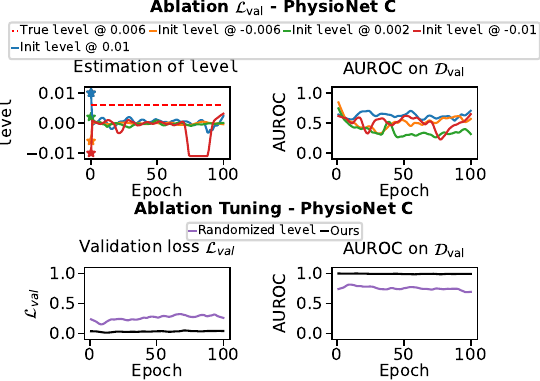}
    }
    \caption{\smash{\textbf{Overview of ablation studies.}} \textbf{Top:} \method's $\Lval$ is replaced by a point-wise val.~loss leading to an erroneous estimation of $a$ (left) and poor performance (right). \textbf{Bottom:} \method's self-tuning module is disabled, $a$ is now randomized. Val.~loss indicates poor alignment, reflected in poor performance.}
    \label{fig:abl}
\end{figure}

In Fig.~\ref{fig:abl} (top), we illustrate the {\tt level} estimation for PhysioNet C under the condition where our Wasserstein-based $\Lval$ is substituted with a point-wise metric, as used in \cite{Yoo23End}. This comparison shows that a point-wise validation loss tends to favor solutions where the {\tt level} of the Trend anomaly approximates zero, neutralizing the anomaly. Although this might produce high alignment, it leads to poor $\fdet$ performance in $\Dval$ (right). This shows that the \textbf{{distributional characteristics captured by our $\Lval$ are a key contributing factor}} to the success of \method.

\paragraph{Randomization vs.~Tuning:~}
\sloppy{
In Fig.~\ref{fig:abl} (bottom), the self-tuning module is disabled for PhysioNet C, where {\tt level} is instead randomized (along with {\tt location}, and {\tt length}).
We observe substantially higher $\Lval$, indicating poor alignment. In turn, $\fdet$ struggles to detect the unlabeled anomalies in $\Dval$. This showcases the \textbf{{utility of \method's systematic hyperparameter (self-) tuning over random choice}}.\looseness=-1
}

\section{Related Work}
\label{sec:related}

\paragraph{SSL for Anomaly Detection:~}

SSL has emerged as a significant approach in machine learning.
Foundation models \cite{bommasani2021opportunities} like large language models \cite{openai2023gpt4,ramesh2021zero}, whose training heavily relies on SSL, have shaken up the world with outstanding performance.
SSL has also transformed representation learning and significantly boosted several tasks in NLP, computer vision, recommender systems and medicine \cite{baevski2022data2vec,krishnan2022self}. 
SSL is especially attractive for unsupervised problems like anomaly detection (AD) \cite{hojjati2022self}, because of its nature to create proxy tasks and loss functions without any labels.
There are various types of SSL-based methods on AD \cite{liu2021self}, but the core idea is to use data augmentation functions \cite{ijcai2021p198,Golan18GEOM,Li21CutPaste} on inliers to create pseudo anomalies, and learn a classifier that can detect the pseudo anomalies.








\vspace{-0.015in}
\paragraph{Time Series Anomaly Detection:~}


There exists a large body of work on TSAD, for which we refer to several surveys \cite{blazquez2021review,gupta2013outlier}.
Recently, SSL has been widely studied as the main approach to address TSAD problems \cite{wen2020time,zhang2023self}.
BeatGAN \cite{zhou2019beatgan} and RobustTAD \cite{gao2020robusttad} use data augmentation to enrich training data.
COUTA \cite{xu2022calibrated} proposes three different augmentations, respectively mimicking local, contextual and collective anomalies.
NeuTraL-AD \cite{qiu2021neutral} propose several augmentation functions that are an integral part of the learning process.
AMSL \cite{zhang2022adaptive} also uses signal transformations as data augmentation.
TimeAutoAD \cite{jiao2022timeautoad} employs three different strategies to augment the training data for generating pseudo anomalies.
Motivated by masked language models \cite{DevlinCLT19}, 
MAD \cite{fu2022mad} investigated various masking procedures for multivariate TSAD based on a predictive pretext task.
Note that only \cite{jiao2022timeautoad,qiu2021neutral} parameterize the augmentation procedures, but they tune those using \textit{labeled} validation data. 


\vspace{-0.015in}
\paragraph{Unsupervised Model Selection:~}
Unsupervised hyperparameter tuning (i.e. model selection) is nontrivial in anomaly detection due to the absence of labeled validation data \cite{ma2023need}, and the literature is slim with quite recent efforts \cite{zhao2022towards,hyper23,MetaOD21,zhao2022toward}.
Specifically on SSL-based AD, a recent study \cite{yoo2023data} has revealed the impact of the choice of augmentation on SSL-based AD, leading to several works that aim to automatically search for the optimal choice \cite{Yoo23End,Yoo23DSV}.
Their main contribution is a novel unsupervised loss, but it is not differentiable.
None of the existing works addressed the model selection problem of SSL-based AD for time series data.

\section{Conclusion}
\label{sec:conclusion}
We introduced \method for self-supervised time series anomaly detection, which is the first attempt that automatically (self-)tunes the augmentation hyperparameters on time series data in an unsupervised manner.
\method includes a differentiable model to augment the input time series data with various anomaly types, and an unsupervised validation loss that assists in aligning the augmented and test data.
Experiments showed \method's ability in effectively selecting the augmentation type along with its continuous hyperparameters. 
Across various real-world datasets with different types of time series anomalies, \method outperformed a diverse list of baselines, including modern neural and self-supervised approaches. 
While being the first self-tuning SSL solution to TSAD, our work opens new research directions. 
For instance, future extensions of \method could include an expanded catalog of supported anomaly types, broadening its applicability. Additionally, \method could be enhanced to deal with multiple different anomaly types within a given dataset, further strengthening its robustness. There is also potential to test the framework on various other datasets from different domains. Moreover, these ideas could be expanded to multivariate time series data, allowing \method to tackle more complex temporal relationships and dependencies effectively.

\clearpage

\section*{Appendix}
All additional materials used in this work are available at the following repository: {\url{https://tinyurl.com/hhfdrrtk}

\bibliographystyle{splncs04}
\bibliography{00ref.bib}

\clearpage
\appendix
\section{Appendix}

\subsection{Types of Time Series Anomalies} \label{ssec:tsatypes}

We include six types of time series anomalies, presented in Table \ref{tab:tsa-types}.
\begin{table}[!b]
    \centering
    \caption{Types of Time Series Anomalies}
    \label{tab:tsa-types}
    \resizebox{\columnwidth}{!}{
    \begin{tabular}{p{2.5cm}|p{7.5cm}}
    \toprule
        \textbf{Type} & \textbf{Description} \\
        \midrule
        Platform & Starting at timestamp {\tt location}, the values of a duration {\tt length} in the time series are \textbf{equal to} a constant value {\tt level}. \\
        Mean shift & Starting at timestamp {\tt location}, a constant value {\tt level} is \textbf{added to} the values of a duration {\tt length} in the time series. \\
        Amplitude & Starting at timestamp {\tt location}, a constant value {\tt level} is \textbf{multiplied with} the values of a duration {\tt length} in the time series. \\
        Trend & Starting at timestamp {\tt location}, a \textbf{series of values $at$ is added to} the duration {\tt length}, where $a$ is the {\tt level} and $t$ is the timestamp in that duration. \\
        Extremum/Spike & A large (either positive or negative) value {\tt level} is \textbf{assigned to} a single timestamp {\tt location} in the time series. \\
        Frequency shift & Starting at phase {\tt location}, the frequency of the duration with {\tt length} phases is \textbf{increased by} a constant value {\tt level}. \\
        \bottomrule
    \end{tabular}
    }
\end{table}
Fig.~\ref{fig:faug} visualizes each type of anomaly and demonstrates how each hyperparameter controls the injection of the anomaly in the time series for a platform anomaly.
%
%
\subsection{Dataset Details}
\label{ssec:datasets}
\paragraph{PhysioNet:~} 
The 2017 PhysioNet Challenge dataset \cite{Clifford2017physionet} comprises a diverse array of real-world 1-lead 300 Hz ECG recordings. 
We use ECG recordings, each 9 seconds in length with $K=2700$ time-steps and standardized to have a zero mean and standard deviation of one. Injected anomalies represent 10\% of the data.
Here, we include three additional controlled TSAD tasks based on the PhysioNet data as shown in Table \ref{tab:additional-datasets}. Table~\ref{table:space} (top) shows the hyperparameter spaces used to train $\faug$ in PhysioNet. The hyperparameters {\tt location} and {\tt length} are normalized by $K$. Noting that the extremum anomaly always occurs on a single timestamp in the time series, thus {\tt length} is always $1$.\looseness=-1

\paragraph{MoCap:~}
The CMU Motion Capture (MoCap) dataset\footnote{http://mocap.cs.cmu.edu/} includes signal data from various sensors on subjects' bodies as they perform different activities (walking, jumping, or running).
As we focus on a univariate setting, only the sensor signal on the left femur is used.
Since each data contains the time series data with a short length, we stitch them into a longer one in our experiment.
To ensure it is done smoothly, we identified the start and end points of each gait phase for the stitching.
We further add random noises to augment the normal samples in the dataset.
Each signal is normalized between $-1$ and $1$ and truncated to length $K=1500$. 
Constructed anomalies represent 10\% of the data. 
As opposed to PhysioNet, we consider MoCap as a dataset with natural or real anomalies. Therefore, we do not create additional TSAD tasks from this dataset. Table~\ref{table:space} (bottom) shows the hyperparameter spaces used to train $\faug$ in MoCap.
The hyperparameters {\tt location} and {\tt length} of platform anomaly are normalized by $K$.
The hyperparameters {\tt location} and {\tt length} of frequency anomaly denote the starting gait phase and the length of gait phases, respectively.\looseness=-1

\begin{table}
\caption{Anomaly profile of \textbf{additional} TSAD tasks.}
\centering
\resizebox{\columnwidth}{!}{
\begin{tabular}{l | ccc}
    \toprule
      Profile & PhysioNet E & PhysioNet F & PhysioNet G \\
      \midrule
      \texttt{Type} & Mean shift & Mean shift & Extremum\\
      \texttt{Level} & Fixed & Fixed & Fixed\\
      \texttt{Location} & Random & Random & Random \\
      \texttt{Length} & Random & Fixed & N/A\\
     \bottomrule
\end{tabular}
}
\label{tab:additional-datasets}
\vspace{-0.3in}
\end{table}


\begin{table}[!t]
\caption{
    Hyperparameter space for each anomaly type.
}
\centering
\resizebox{1\columnwidth}{!}{
\begin{tabular}{l | ccc }
\toprule
\multicolumn{4}{c}{PhysioNet} \\
\midrule
\textbf{Anomaly Types} & \tt{location} & \tt{length} & \tt{level} \\
\midrule
Platform & $[100..2000]$ & $[400..600]$ & $\{0.2k-1 | k \in [0..10]\}$ \\
Mean Shift & $[100..2000]$ & $[400..600]$ & $\{0.2k-1 | k \in [0..10]\}$ \\
Amplitude & $[100..2000]$ & $[400..600]$ & $\{0.5k+1 | k \in [0..10]\}$ \\
Trend & $[100..2000]$ & $[400..600]$ & $\{0.002k-0.01 | k \in [0..10]\}$ \\
Extremum/Spike & $\{100k | k \in [1..26]\}$ & $\{1\}$ & $\{3k-15 | k \in [0..10]\}$ \\
\midrule
\multicolumn{4}{c}{MoCap} \\
\midrule
\textbf{Anomaly Types} & {\tt location} & {\tt length} & {\tt level} \\
\midrule
Platform & $[200..800]$ & $[100..200]$ & $\{0.2k-1 | k \in [0..10]\}$ \\
Frequency Shift & $[1..3]$ & $[1..6]$ & $[1..3]$ \\
\bottomrule
\end{tabular}}
\label{table:space}
\end{table}

\subsection{Model Configurations}
\label{ssec:config}

\noindent See below for details on model configurations (cf. Sec. \ref{sec:exp_setup}).

\paragraph{\method configuration:~}
In Table \ref{tab:model_config}, we provide a comprehensive overview of the configuration details for the different components of \method. 

\begin{table}[!b]
\centering
\caption{Configuration details for $\faug$ and $\fdet$ of \method}
\label{tab:model_config}
\resizebox{\columnwidth}{!}{
\begin{tabular}{l | l}
\toprule
\multicolumn{2}{c}{\textbf{$\faug$ configuration}} \\
\midrule
\textit{Encoder$_{\bphi}$} & \\
\quad Conv Layer 1 & \{In: 1, Out: 64, Kernel: 100, Stride: 4, ReLU, BatchNorm\} \\
\quad Conv Layer 2 & \{In: 64, Out: 64, Kernel: 100, Stride: 4, ReLU\} \\
\textit{Decoder$_{\bphi}$} & \\
\quad TransConv Layer 1 & \{In: 64, Out: 64, Kernel: 100, Stride: 4, ReLU, BatchNorm\} \\
\quad TransConv Layer 2 & \{In: 64, Out: 1, Kernel: 100, Stride: 4\} \\
\textit{MLP$_{\bphi}$} & \\
\quad MLP Layer 1 &  \{In: 3, Out: 16, ReLU\} \\
\quad MLP Layer 2 &  \{In: 16, Out: dim$(\Ztrn)$, ReLU\} \\
\textit{General Parameters} & \\
\quad Batch Size & 64 \\
\quad \# Epochs & 500 \\
\quad Optimizer & Adam (LR: 0.002) \\
\midrule
\multicolumn{2}{c}{\textbf{$\fdet$ configuration}} \\
\midrule
\textit{Encoder$_{\btheta}$} & \\
\quad Conv Layer 1 & \{In: 1, Out: 32, Kernel: 10, Stride: 2, ReLU, BatchNorm\} \\
\quad Conv Layer 2 & \{In: 32, Out: 16, Kernel: 10, Dilation: 2, Stride: 2\} \\
\quad Conv Layer 3 & \{In: 16, Out: 8, Kernel: 10, Dilation: 4, Stride: 4\} \\
\quad Avg Pooling + Flatten & \{Kernel: 10, Stride: 3\} \\
\quad Linear Layer & \{In: 400, Out: 10\} \\
\textit{MLP$_{\btheta}$} & \\
\quad MLP Layer 1 & \{In: 10, Out: 1\} \\
\textit{General Parameters} & \\
\quad Dropout & 0.2 \\
\quad Batch Size & 64 \\
\quad Warm Start \# Epochs $\fdet$ & 3 \\
\quad \# Epochs & 100 \\
\quad Optimizer for $\ba$ & Adam (LR: 0.001) \\
\quad Optimizer for $\fdet$ & Adam (LR: 0.002) \\
\quad Mixing Rate (cf. Phase ii) & 0.15 \\
\bottomrule
\end{tabular}
}
\vspace{-0.3in}
\end{table}

\paragraph{Baseline configurations:~}
The details for the baseline configurations are provided in Table \ref{tab:baselines}.
\\
Model training was done on a NVIDIA Tesla P100 GPU.

\begin{table}[!h]
    \centering
    \vspace{-0.2in}
    \caption{Configuration Details for Baselines}
    \label{tab:baselines}
    \resizebox{\columnwidth}{!}{
    \begin{tabular}{p{2cm}|p{8cm}}
    \toprule
        \textbf{Method} & \textbf{Hyperparameter Settings} \\
        \midrule
        OC-SVM & We use author-recommended hyperparameters \cite{Hsu03svm}. \\
        LOF & We use author-recommended hyperparameters \cite{Breunig00LOF}. \\
        IF & We use author-recommended hyperparameters \cite{Liu08IF}. \\
        ARIMA & We use AutoARIMA to select hyperparameters \cite{box2015time}. \\
        MP & We use author-recommendations to set the window size $m$ \cite{yeh2016mp}. \\
        EncDec-LSTM & We downsample our time series to length approx. 200, following \cite{malhotra2016lstm}. Other hyperparameters follow authors' recommendations. \\
        SR-CNN & We use author-recommended hyperparameters \cite{Hansheng19SRCNN}. \\
        USAD & We downsample our time series to length approx. 200, following \cite{Audibert20USAD}. Other hyperparameters follow authors' recommendations. \\
        NeuTraL-AD & We use author-recommended hyperparameters \cite{qiu2021neutral}. We tune augmentation type for each dataset using \textit{labeled} validation data. \\
        TimeGPT & We tune the confidence interval \cite{garza2023timegpt} for each dataset using \textit{labeled} validation data. \\
        \bottomrule
    \end{tabular}

    }
\end{table}

\vspace{-0.2in}
\subsection{Additional Results}
\label{ssec:addresults}
We present additional results for continuous and discrete augmentation hyperparameter tuning and two additional ablation studies.
\subsubsection{Cont. Aug. Hyperparameter Tuning}
In addition to the results on continuous hyperparameter tuning for PhysioNet A and B (see Fig.~\ref{fig:platform-estimation}), we demonstrate \method's efficacy in tuning the continuous hyperparameters on five additional TSAD tasks. These include PhysioNet C and D, which feature Trend anomalies (cf. Table \ref{tab:datasets}), as well as PhysioNet E and F, showcasing Mean shift anomalies, and PhysioNet G, illustrating Extremum anomalies (cf. Table \ref{tab:additional-datasets}).

\paragraph{PhysioNet C \& D:~} 

\begin{figure}[!b]
    \centering
    \resizebox{\columnwidth}{!}{%
    \includegraphics[scale=0.92]{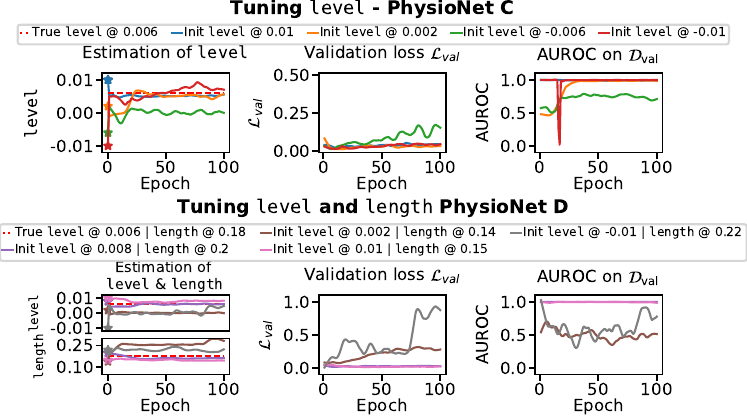}
    }
    \vspace{-0.2in}
    \caption{\textbf{Tuning the continuous augmentation hyperparameter(s)  
 with \method for Trend anomalies}. \textbf{Top:} Given Trend anomalies at true {\tt level} (red dashed line), various initializations converge near the true value (left), following the minimized values of val. loss (center), and leading to high detection AUROC performance (right). \textbf{Bottom:} Multiple continuous hyperparameters, here both {\tt level} and {\tt length} are tuned to near true values (left), guided by minimizing the val. loss (center), achieving high AUROC (right).}
 
    \label{fig:trend-estimation}
\end{figure}

The tuning process of the continuous hyperparameters for the \textbf{Trend anomalies} in PhysioNet C and D is shown in Fig. \ref{fig:trend-estimation}. We observe several initializations for $\ba$ that arrive closely to the true {\tt level} for PhysioNet C, as well as to the true {\tt level} and {\tt length} for PhysioNet D. In turn, those initializations yield a detector $\fdet$ with high performance on $\Dval$. Note how, for PhysioNet C, $\Lval$ is low across the board. This is likely due to the fact that a Trend anomaly with a subtle slope, has similar characteristics to inliers (see e.g. Fig. \ref{fig:mismatch} bottom left). Yet, \method effectively assigns a higher validation loss to the initializations that lead to misaligned cases. This shows the effectiveness of our method even in cases where anomalies are subtle.

\paragraph{PhysioNet E \& F:~}

Similarly to Trend anomalies, \textbf{Mean shift anomalies} are inherently subtle, especially when the {\tt level} is close to zero. We show in Fig.~\ref{fig:mean-estimation} how \method properly tunes the continuous hyperparameters for {\tt level} and {\tt length} in PhysioNet E and F for several initializations.

\paragraph{PhysioNet G:~}

Lastly, Fig. \ref{fig:spike-estimation} showcases \method's ability to tune the {\tt level} of the spike in the \textbf{Extremum anomalies} while the location is randomized. Note that Extremum anomalies have no {\tt length} by definition. The ECG recordings in PhysioNet contain many natural spikes. As such, validation loss is low by default. Nonetheless, \method successfully tunes two out of four initializations and reflects -- though subtly -- this difference in the validation loss. This again leads to a well-tuned $\fdet(\cdot;\btheta^*)$ that performs strongly on $\Dval$. 

\begin{figure}[!b]
    \centering
    \resizebox{\columnwidth}{!}{
    \includegraphics[scale=0.92]{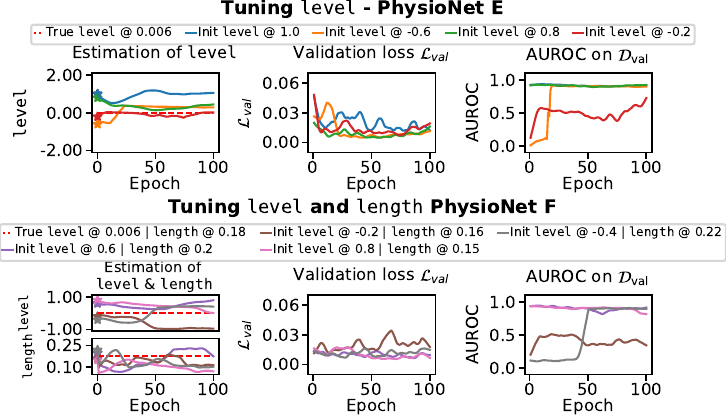}
    }
    \vspace{-0.1in}
    \caption{\textbf{Tuning the continuous augmentation hyperparameter(s)  
 with \method for Mean shift anomalies}. 
 \textbf{Top:} Given Mean shift anomalies at true {\tt level} (red dashed line), various initializations converge near the true value (left), following the minimized values of val. loss (center), and leading to high detection AUROC performance (right). \textbf{Bottom:} Multiple continuous hyperparameters, both {\tt level} and {\tt length} are accurately tuned to near true values (left), as guided by minimizing the val. loss (center), achieving high AUROC (right). Mean shift anomalies have low val. loss by default due to the subtle nature of Mean shift anomalies.\looseness=-1
 }
    \label{fig:mean-estimation}
\end{figure}

\vspace{-0.1in}
\subsubsection{Discrete Augmentation Hyperparameter Tuning}
We showcased \method's ability to tune the discrete hyperparameter, anomaly type, in Fig.~\ref{fig:mismatch} for controlled and natural TSAD tasks. Given the direct applicability and significance of discrete hyperparameter tuning in real-world contexts, we present extended results for discrete hyperparameter tuning. 

\paragraph{MoCap B:~} Fig.~\ref{fig:discrete-mocapB} shows the\textbf{ discrete hyperparameter tuning for the unknown anomaly type} in MoCap B. \method was initialized twice: first with $\faug$ pre-trained for injecting Frequency shift anomalies, and second with $\faug$ pre-trained for injecting Platform anomalies. The validation loss (left) indicates a strong alignment between $\Dtrn \cup \Daug$ and the unlabeled $\Dval$ when \method is initialized with Frequency shift anomalies. This is also reflected in $\fdet$'s performance on $\Dval$ (center). Visually, we can indeed confirm that Frequency shift anomalies (right -- red) appear to be more similar to the running pattern (right -- black) as opposed to Platform anomalies (right -- purple).

\begin{figure}[!b]
    \centering
    \resizebox{\columnwidth}{!}{    
    \includegraphics[scale=0.96]{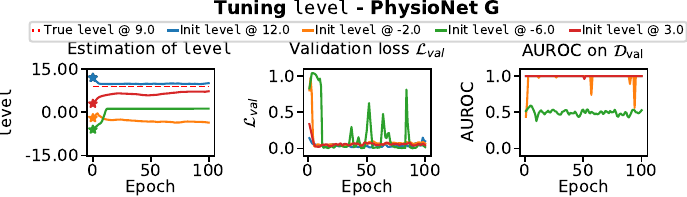}
    }
    \vspace{-0.1in}
    \caption{\textbf{Tuning the continuous augmentation hyperparameter(s)  
 with \method for Extremum anomalies}. 
 Given Extremum anomalies at true {\tt level} (red dashed line), various initializations converge near the true value (left), following the minimized values of val. loss (center), and leading to high detection AUROC performance (right). Note how Extremum anomalies have low val. loss by default due to the natural presence of spikes in ECG data.
 }
    \label{fig:spike-estimation}
\end{figure}

\subsubsection{Additional Ablation Studies}

Following Sec.~\ref{ssec:abl}, we present additional ablation studies for the second-order optimization within \method and the normalization of the embeddings $\{\Ztrn, \Zaug, \Zval\}$ obtained through $\fdet^\mathrm{enc}$.
\paragraph{Second-order Opt. Ablation:~}

Fig.~\ref{fig:abl-extra} (top) shows the {\tt level}-estimation and performance of $\fdet$ on PhysioNet C when the second-order optimization is disabled. Note how the estimation process becomes highly unstable when second-order optimization is disabled. In turn, performance of $\fdet$ on $\Dval$ suffers severely.

\paragraph{Embedding Normalization Ablation:~}

We show the {\tt level}-estimation and performance of $\fdet$ on PhysioNet C when embedding normalization is disabled in Fig.~\ref{fig:abl-extra} (bottom). While $a$ initialized at $0.01$ eventually leads to the correct {\tt level}, the estimation process is highly volatile compared to when normalization is enabled as shown in Fig.~\ref{fig:trend-estimation} (top).

\begin{figure}[!h]
    \centering
    \resizebox{\columnwidth}{!}{
    \includegraphics{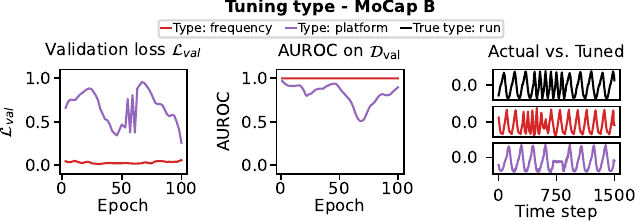}
    }
    \vspace{-0.1in}
    \caption{\textbf{Tuning discrete hyperparameter (anomaly type) with \method}. 
    For Run anomalies in MoCap B with unknown type (black), val. loss favors Frequency shift (red) that leads to high AUROC (center) and mimics well true anomaly (right), and effectively rejects inferior type Platform (purple).
    }
    \label{fig:discrete-mocapB}
\end{figure}

\begin{figure}[!h]
    \centering
    \resizebox{\columnwidth}{!}{
    \includegraphics[scale=0.5]{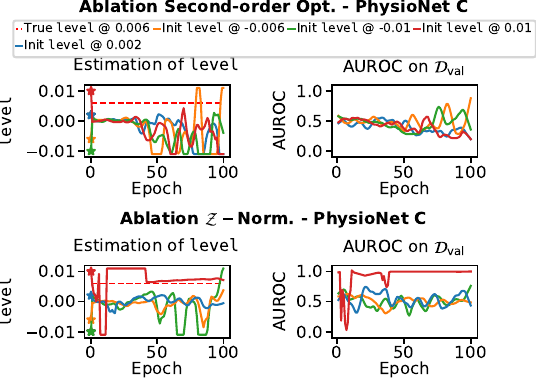}
    }
    \vspace{-0.1in}
    \caption{\textbf{Overview of additional ablation studies.} \textbf{Top:} 
    We \textbf{disable second-order optimization} in \method, leading to a highly unstable estimation process of $a$ (left) and poor performance of $\fdet$ (right). \textbf{Bottom:} We \textbf{disable normalization} of the embeddings in \method. Estimation of $a$ (left) is volatile and does not converge well, in turn, performance of $\fdet$ on $\Dval$ is poor in most cases (right).
    }
    \label{fig:abl-extra}
\end{figure}

\end{document}